\documentclass{article}

\usepackage{microtype}
\usepackage{graphicx}
\usepackage{subcaption}
\usepackage{booktabs} 
\usepackage{multirow}
\usepackage{hyperref}

\usepackage[preprint]{icml2026}

\usepackage{amsmath}
\usepackage{amssymb}
\usepackage{mathtools}
\usepackage{amsthm}

\usepackage{xcolor}
\usepackage{kotex}

\usepackage[capitalize,noabbrev]{cleveref}

\theoremstyle{plain}

\theoremstyle{definition}

\theoremstyle{remark}

\usepackage[textsize=tiny]{todonotes}

\icmltitlerunning{Generalizing Code-Switching ASR to Unseen Language Pairs}

\begin{document}

\twocolumn[
  \icmltitle{Towards Truly Multilingual ASR:\\
    Generalizing Code-Switching ASR to Unseen Language Pairs}


  \icmlsetsymbol{equal}{*}

  \begin{icmlauthorlist}
    \icmlauthor{Gio Paik}{equal,thetaone,kitsch}
    \icmlauthor{Hyunseo Shin}{equal,uos}
    \icmlauthor{Soungmin Lee}{kitsch,gtech}
  \end{icmlauthorlist}

  \icmlaffiliation{thetaone}{Theta One Korea, Seoul, Republic of Korea}
  \icmlaffiliation{uos}{Department of Artificial Intelligence, University of Seoul, Seoul, Republic of Korea}
  \icmlaffiliation{gtech}{College of Computing, Georgia Institute of Technology, GA, USA}
  \icmlaffiliation{kitsch}{Kitsch Labs, Seongnam, Republic of Korea}

  \icmlcorrespondingauthor{Gio Paik}{giopaik0@gmail.com}

  \icmlkeywords{ICML, Speech Recognition, ASR, Code-Switching, Generalization, Model Merging}

  \vskip 0.3in
]



\printAffiliationsAndNotice{\icmlEqualContribution}

\begin{abstract}
Automatic Speech Recognition (ASR) has become a key technology for human--AI interaction. However, code-switching ASR (CS-ASR) remains particularly challenging due to the severe scarcity of multilingual CS speech resources across diverse language pairs. Existing approaches primarily improve CS-ASR performance through synthetic CS speech generation or pair-specific fine-tuning on limited bilingual datasets. Nevertheless, these approaches face an inherent scalability limitation, as support for CS must be developed separately for language pairs whose number grows combinatorially with the number of supported languages. 
In this work, we investigate whether CS capabilities learned from a limited set of seen language pairs can generalize to unseen language pairs through model merging and domain generalization methods.
Our experiments show that merged bilingual CS-ASR models modestly generalize to unseen language pairs, suggesting limited transfer of bilingual CS capabilities across language pairs.
\end{abstract}

\section{Introduction}
Despite advances in Automatic Speech Recognition (ASR), code-switching (CS)---the alternation of multiple languages within a single utterance---remains a significant challenge for ASR systems. The core issue lies in the scarcity of CS speech data: for a multilingual ASR model supporting $N$ languages, the number of possible language pairs grows quadratically with $N$, making it practically infeasible to collect CS speech data for every pair. To address this limitation and enable robust recognition of CS speech commonly used by multilingual speakers, it is necessary to develop ASR systems that can generalize CS capabilities to all $\binom{N}{2}$ language pairs using training data from only a limited subset of language pairs.

In this paper, we investigate whether CS capabilities learned from a subset of language pairs can generalize to unseen language pairs, using speech data from four languages: English (\textsc{en}), Korean (\textsc{ko}), Japanese (\textsc{ja}), and German (\textsc{de}). Specifically, we examine whether CS capabilities acquired from relatively accessible language pairs--\textsc{ko-en}, \textsc{ja-en}, and \textsc{de-en}--can transfer to unseen pairs such as \textsc{ko-ja} and \textsc{ko-de}. To this end, we explore model merging and domain generalization, and construct small-scale \textsc{ko-ja}\footnote{The \textsc{ko-ja} evaluation dataset is available at \url{https://huggingface.co/datasets/thetaone-ai/Korean-Japanese-Code-Switching-Speech}.} and \textsc{ko-de} CS-ASR evaluation datasets.

Our results show that fine-tuning on one language pair yields slight gains on other pairs, while model merging and domain generalization (DG) can further improve recognition on unseen language pairs. However, the gains remain limited, highlighting the need for methods tailored to the characteristics of CS-ASR, rather than a naive application of existing model merging or domain generalization techniques.

Our contributions are threefold:
\begin{enumerate}
  \item We systematically investigate  whether CS-ASR capabilities learned from specific language pairs can generalize to unseen language pairs.
  \item We show the limitations of directly applying existing model merging and domain generalization methods to CS-ASR.
  \item We construct the first Korean-Japanese and Korean-German CS speech evaluation datasets, and will open-source the Korean-Japanese dataset.
\end{enumerate}

\section{Related Works}
\subsection{Code-Switching Speech Recognition and Datasets}
Code-switching ASR research has largely focused on Chinese-English \cite{ASRU, CS-Dialogue, DOTA-ME-CS, talcs}, with comparatively limited coverage of other English-centric pairs such as English-Korean \cite{hike} and English-Hindi \cite{dey-fung-2014-hindi}. In contrast, publicly available datasets for non-English language pairs, such as Korean-Japanese or Korean-German, remain virtually nonexistent.
To address this limitation, recent studies have attempted to synthesize code-switching speech using TTS systems \cite{cs-fleurs-not-unicom, yu2023code, Sharma2020ImprovingLR} or by concatenating monolingual speech segments \cite{unicom}. However, such approaches often generate acoustically unnatural code-switching speech due to limited CS-aware synthesis capability.


Consequently, most CS-ASR studies have focused on improving recognition performance for individual language pairs. Prior work explored bilingual linguistic biases \cite{chi-bell-2022-improving,10448335}, language-specialized architectures \cite{kulkarni2023adapting, zhang2025boosting}, and CS text-based adaptation methods \cite{nguyen2025asyncswitch, pandey2025whistle}. However, existing approaches still primarily target seen language pairs rather than generalizing code-switching capabilities to unseen pairs.

\subsection{Model Merging}
Model merging has recently emerged as an efficient alternative to multi-task retraining for combining independently fine-tuned models. Early approaches such as Task Arithmetic \cite{ilharco2023editing} demonstrated that task-specific parameter differences can be linearly combined in weight space to transfer or compose capabilities across tasks. Subsequent studies further explored more robust merging strategies including TIES-Merging \cite{yadav2023ties}, which resolves parameter conflicts through sparse sign agreement, and DARE \cite{yu2024language}, which improves merge robustness through random pruning and rescaling.

Recent work has shown that model merging can effectively combine capabilities across diverse low-resource domains, including multilingual language modeling \cite{tao-etal-2024-unlocking, bandarkar2025layer, shin-hwang-2026-layer} and multimodal vision-language models \cite{pmlr-v267-chen25cm, wei2026optmerge}. 
In ASR, \citet{ducorroy2025robust} used model merging to improve robustness to out-of-distribution speech, including disordered speech, while \citet{rolland2025group} applied model merging to child ASR.
However, its application to multilingual CS-ASR, remains unexplored.

\subsection{Domain Generalization}
Domain generalization has become an important research direction for learning robust representations across heterogeneous domains, particularly in computer vision.
Recent work reformulate DG from an optimization perspective, using gradient consistency across domains: MLDG \cite{li2018learning} applies meta-learning to optimize updates for unseen domains, Fish \cite{shi2021gradient} maximizes gradient agreement, Fishr \cite{rame2022fishr} aligns domain-level gradient variances to regularize loss landscapes, and Gradient-Guided Annealing (GGA) \cite{ballas2025gradient} mitigates domain overfitting via early-stage gradient alignment.
However, most existing DG methods have been explored primarily in computer vision, with relatively limited investigation in low-resource ASR settings.

\begin{table*}[t]
  \caption{Mixed Error Rate (MER) on the dataset for each code-switching language pair. Lower is better.}
  \label{table:main_result}
  \begin{center}
    \begin{small}
      \begin{sc}
        \begin{tabular}{l|ccc|c|cc|c}
          \toprule
          & \multicolumn{4}{c|}{\textbf{Seen}} & \multicolumn{3}{c}{\textbf{Unseen}} \\
          & ko-en & ja-en & de-en & Avg. & ko-de & ko-ja & Avg. \\
          \midrule
            
          Whisper-medium 
          & 0.26 & 0.56 & 0.15 & 0.33 & 0.39 & 0.44 & 0.41 \\
          \midrule
          ko-en FT
          & 0.12 & 0.23 & 0.12 & 0.16 & 0.35 & 0.46 & 0.40 \\
          ja-en FT
          & 0.14 & 0.28 & 0.13 & 0.18  & 0.38 & 0.31 & 0.35 \\
          de-en FT
          & 0.14 & 0.31 & 0.12 & 0.19 & 0.38 & 0.35 & 0.36 \\
          ko-en + ja-en + de-en FT
          & 0.11 & 0.38 & 0.12 & 0.20 & 0.40 & 0.41 & 0.41 \\
          \midrule
          
          \multicolumn{8}{c}{\textbf{Model Merging}} \\
          \midrule
          \textbf{Task Arithmetic \cite{ilharco2023editing}} & & & & & & & \\
          \quad ko-en + ja-en
          & 0.20 & 0.24 & 0.18 & 0.20 & 0.36 & 0.53 & 0.45 \\ 
          \quad ko-en + de-en 
          & 0.12 & 0.29 & 0.16 & 0.19 & 0.36 & 0.40 & 0.38 \\ 
          \quad ja-en + de-en 
          & 0.17 & 0.24 & 0.15 & 0.19 & 0.47 & 0.47 & 0.47 \\ 
          \quad ko-en + ja-en + de-en 
          & 0.73 & 0.61 & 0.34 & 0.56 & 0.57 & 0.96 & 0.77 \\
          \midrule

          \textbf{TIES \cite{yadav2023ties}} & & & & & & & \\
          \quad ko-en + ja-en
          & 0.11 & 0.20 & 0.12 & 0.14 & 0.34 & 0.31 & 0.32 \\ 
          \quad ko-en + de-en 
          & 0.11 & 0.21 & 0.12 & 0.15 & 0.37 & 0.39 & 0.38 \\ 
          \quad ja-en + de-en 
          & 0.12 & 0.25 & 0.12 & 0.16 & 0.44 & 0.36 & 0.40 \\ 
          \quad ko-en + ja-en + de-en 
          & 0.11 & 0.20 & 0.11 & 0.14 & 0.37 & 0.30 & 0.34 \\
          \midrule

          \textbf{DARE \cite{yu2024language}} & & & & & & & \\
          \quad ko-en + ja-en
          & 0.21 & 0.24 & 0.19 & 0.21 & 0.37 & 0.57 & 0.47 \\ 
          \quad ko-en + de-en 
          & 0.12 & 0.28 & 0.16 & 0.19 & 0.36 & 0.40 & 0.38 \\ 
          \quad ja-en + de-en 
          & 0.16 & 0.28 & 0.15 & 0.20 & 0.48 & 0.47 & 0.48 \\ 
          \quad ko-en + ja-en + de-en 
          & 0.74 & 0.58 & 0.34 & 0.55 & 0.58 & 0.96 & 0.77 \\
          \midrule
          
          \multicolumn{8}{c}{\textbf{Domain Generalization}} \\
          \midrule
          Fish~\cite{shi2021gradient}
          & 0.11 & 0.25 & 0.15 & 0.17 & 0.47 & 0.53 & 0.50 \\
          Fishr~\cite{rame2022fishr}
          & 0.11 & 0.29 & 0.13 & 0.18 & 0.35 & 0.31 & 0.33 \\
          GGA-L~\cite{ballas2025gradient}
          & 0.11 & 0.28 & 0.13 & 0.17 & 0.45 & 0.40 & 0.42 \\
          \bottomrule  
        \end{tabular}
      \end{sc}
    \end{small}
  \end{center}
  \vskip -0.1in
\end{table*}

\section{Experiments}
\subsection{Training Setup}
We use the widely adopted multilingual ASR model \textsc{Whisper-medium} \cite{whisper} as our backbone and investigate whether code-switching capabilities learned from seen English-centric pairs can improve recognition on unseen non-English-centric pairs.

For the seen language pairs, we fine-tune on three bilingual code-switching datasets, \citet{aihub_ko_en_cs_data} for \textsc{ko-en}, \citet{jecs_ja_en_cs_data} for \textsc{ja-en}, and the \textsc{de-en} split of \citet{unicom}, and evaluate on the human-recorded READ split from \citet{cs-fleurs-not-unicom}.
Since no publicly available datasets exist for the unseen \textsc{ko-ja} and \textsc{ko-de} code-switching pairs, we construct our own evaluation sets. For \textsc{ko-ja}, we collect $450$ code-switching utterances whose scripts are written, recorded, and verified by authors proficient in both Korean and Japanese. For \textsc{ko-de}, we translate the English segments of the \textsc{ko-en} code-switching dataset from \citet{hike} into German using GPT-5.4 mini \cite{openai_gpt54}, and ask two graduate students proficient in Korean and German to review and record the translated utterances, resulting in $387$ speech samples.

Following prior work on multilingual code-switching ASR~\cite{ASRU, CS-Dialogue, hike}, we use \textbf{Mixed Error Rate (MER)}, which accounts for language-specific transcription characteristics within a single utterance. Additional experimental details are provided in Appendix~\ref{appendix:details}.

\subsection{Fine-Tuning with CS Dataset}
We first examine a simple fine-tuning baseline, where \textsc{Whisper-medium} is fine-tuned on the code-switching speech data from a single seen language pair. The top of Table~\ref{table:main_result} shows the performance of the pretrained \textsc{Whisper-medium} model and its variants fine-tuned on each language-pair dataset. Overall, fine-tuning on one CS dataset improves recognition not only on the corresponding language pair but also, to some extent, on other code-switching pairs. This trend is particularly pronounced for \textsc{ja-en}, where the pretrained baseline exhibits a substantially higher MER and all fine-tuning configurations yield clear improvements. However, except for \textsc{de-en}, where the pretrained model already performs relatively well, fine-tuning on a different language pair does not consistently produce large MER reductions, suggesting that naive pair-specific adaptation alone provides limited cross-pair generalization.

\subsection{Merging}
To examine whether model merging can generalize CS-ASR capabilities acquired through fine-tuning on seen language pairs, we merge models fine-tuned on \textsc{ko-en}, \textsc{ja-en}, and \textsc{de-en} using three methods: Task Arithmetic~\cite{ilharco2023editing}, TIES~\cite{yadav2023ties}, and DARE~\cite{yu2024language}.

Among the merging approaches, TIES consistently demonstrates the most stable behavior across all pairwise merge settings. In particular, the TIES merge of \textsc{ko-en} and \textsc{ja-en} models achieves an average MER of $0.14$ on the seen bilingual tasks while maintaining competitive performance on unseen language pairs. Similar trends are observed for the \textsc{ko-en} + \textsc{de-en} and \textsc{ja-en} + \textsc{de-en} settings, suggesting that conflict-aware sparse parameter merging can effectively combine language-pair-specific code-switching capabilities without severe interference.

In contrast, Task Arithmetic and DARE exhibit substantial instability, especially in the three-model merge setting. While pairwise merging remains partially effective, directly combining multiple bilingual CS-ASR models through naive parameter arithmetic often leads to severe degradation.

\subsection{Domain Generalization}
We also evaluate three domain generalization methods, Fish \cite{shi2021gradient}, Fishr \cite{rame2022fishr}, and GGA \cite{ballas2025gradient}, by training on the seen language pairs and measuring their performance on unseen language pairs. For GGA, we use GGA-L, a computationally cheaper variant reported to achieve performance comparable to the full GGA method.

Overall, DG-based fine-tuning does not yield meaningful improvements in MER on unseen pairs, with the exception of Fishr. Fishr improves the average MER on unseen pairs by $0.08$ compared with fine-tuning on data from all seen language pairs. However, its absolute MER remains above $0.3$, indicating that the improvement is still insufficient for robust unseen pair code-switching recognition.

We hypothesize that this limited gain stems from a mismatch between the assumptions of conventional DG methods and the nature of CS-ASR across language pairs. Standard DG methods typically assume that task-relevant mechanisms are shared across domains while domain-specific variations change. In contrast, code-switching across different language pairs changes not only the domain but also the output distribution itself, as the target language composition varies across pairs. As a result, naively applying general-purpose DG methods may be insufficient to achieve substantial generalization to unseen code-switching language pairs.

\section{Fine-Tuned Parameter Analysis}
\begin{figure}[ht]
  \vskip 0.2in
  \begin{center}
    \centerline{\includegraphics[width=0.9\linewidth]{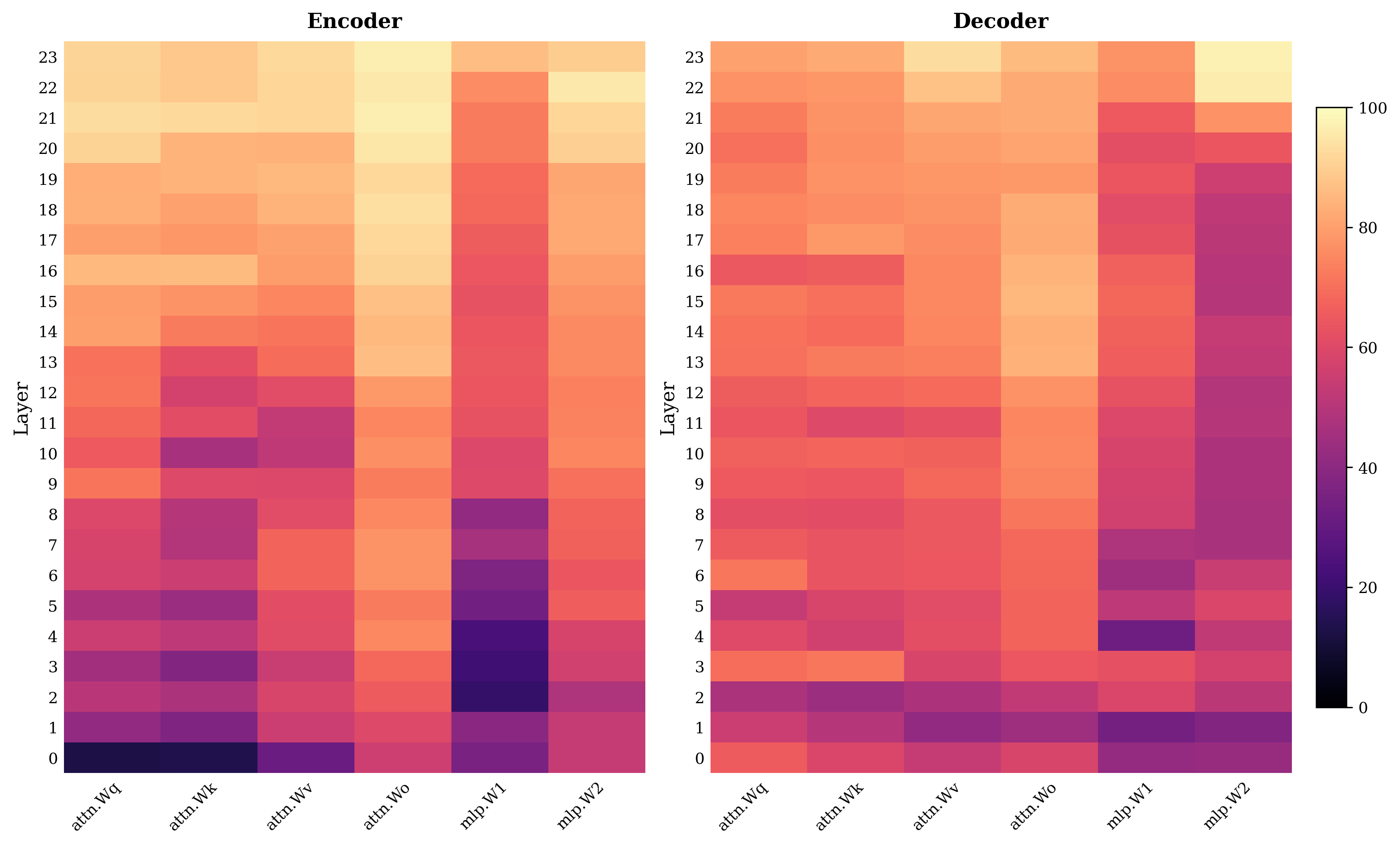}}
    \caption{
      Layer-wise row-level MAV threshold ratios between the pretrained \textsc{Whisper-medium} model and the \textsc{ko-en} code-switching fine-tuned model. Each value represents the percentage of rows whose parameter delta MAV exceeds the predefined threshold.
    }
    \label{fig:mav_ko_en}
  \end{center}
\end{figure}

Figure~\ref{fig:mav_ko_en} visualizes the layer-wise row-level Mean Absolute Value (MAV) ratio of parameter deltas between the pretrained \textsc{Whisper-medium} model and the \textsc{ko-en} code-switching fine-tuned model following \citet{bandarkar2025layer}. We measure the percentage of rows whose delta MAV exceeds a predefined threshold ($5 \times 10^{-5}$) for each projection matrix in the encoder and decoder layers. Thus, higher values indicate that a larger portion of parameters in the corresponding module were substantially updated during code-switching fine-tuning.

Both the encoder and decoder exhibit progressively larger parameter modifications in higher layers, while lower layers remain relatively stable. This trend suggests that code-switching adaptation primarily occurs in deeper semantic and linguistic representations rather than low-level acoustic processing.
The corresponding visualizations for the remaining language pairs are provided in the Appendix~\ref{appendix:parameter}.

\section{Limitations}
First, performance on unseen language pairs remains limited. Although model merging and domain generalization reduce the average MER on unseen pairs to $0.32$, the performance is still far from practical deployment, particularly compared to the sub-$0.2$ MER achieved after fine-tuning on seen pairs. 

Second, both the quantity and diversity of the training and evaluation data are limited. The \textsc{ja-en} training set contains only $582$ utterances from a single speaker, while the \textsc{ko-ja} and \textsc{ko-de} evaluation sets contain recordings from only two speakers per pair, restricting linguistic and speaker diversity. In addition, our unseen pair experiments involve only combinations of languages already observed during training, and therefore do not evaluate generalization to entirely unseen languages such as French or Chinese. These limitations highlight the need for broader multilingual CS-ASR benchmarks and higher-quality CS speech resources.

Third, our experiments are limited to \textsc{Whisper-medium}. A more comprehensive understanding of code-switching generalization will require evaluation across larger Whisper variants and recent audio language models.

Future work should focus on both improving multilingual CS data and developing methods specifically designed for code-switching generalization. Promising directions include analyzing the model components responsible for code-switching behavior, designing domain generalization objectives that explicitly model language-pair shifts, and expanding training and evaluation resources to more diverse language pairs. We view this work as an initial step toward reducing the need to collect code-switching speech data for every possible language pair.

\section{Conclusion}
In this paper, we investigate whether CS-ASR capabilities learned from a limited set of language pairs can generalize to unseen pairs without requiring pair-specific code-switching data. Using \textsc{Whisper-medium} as the backbone, we evaluate fine-tuning and model merging across multilingual code-switching settings involving English, Korean, Japanese, and German.

Our results show that bilingual CS-ASR fine-tuning partially transfers to unseen language pairs, while existing model merging and domain generalization methods remain insufficient to fully bridge the performance gap between seen and unseen pairs. Furthermore, our layer-wise MAV analysis reveals that code-switching adaptation is concentrated in higher encoder and decoder layers, suggesting that generalization to unseen language pairs requires complex task-level adaptations beyond simple domain-level transfer.

These findings highlight the limitations of existing CS-ASR generalization methods and suggest that robust CS-ASR will require architectures and adaptation strategies specifically designed for transferable code-switching capability.

\section*{Acknowledgements}
This work was supported by the Tech Incubator Program for Startup Korea (RS-2024-00507331) funded by the Ministry of SMEs and Startups (MSS, S. Korea).

\bibliography{example_paper}
\bibliographystyle{icml2026}

\newpage
\appendix
\section{Experimental Details}
\label{appendix:details}
\paragraph{Training}
We adopt \textsc{Whisper-medium} \cite{whisper} as the backbone model. All fine-tuning experiments on a single language pair are performed with a batch size of $8$ for $73$ training steps. For \textsc{ko-en} + \textsc{ja-en} + \textsc{de-en} FT and domain generalization experiments, we use a batch size of $9$ for $195$ training steps. We employ the AdamW optimizer with a cosine learning rate decay schedule and a linear warmup phase corresponding to $10\%$ of the total training steps. For model merging methods, including Task Arithmetic, TIES, and DARE, we use MergeKit \cite{mergekit}. All experiments are conducted using PyTorch 2.8.0 on NVIDIA GeForce RTX 4090 GPUs.

\onecolumn
\section{Parameter Analysis}
\label{appendix:parameter}
\begin{figure*}[ht]
  \vskip 0.2in
  \begin{center}
    \centerline{\includegraphics[width=0.9\linewidth]{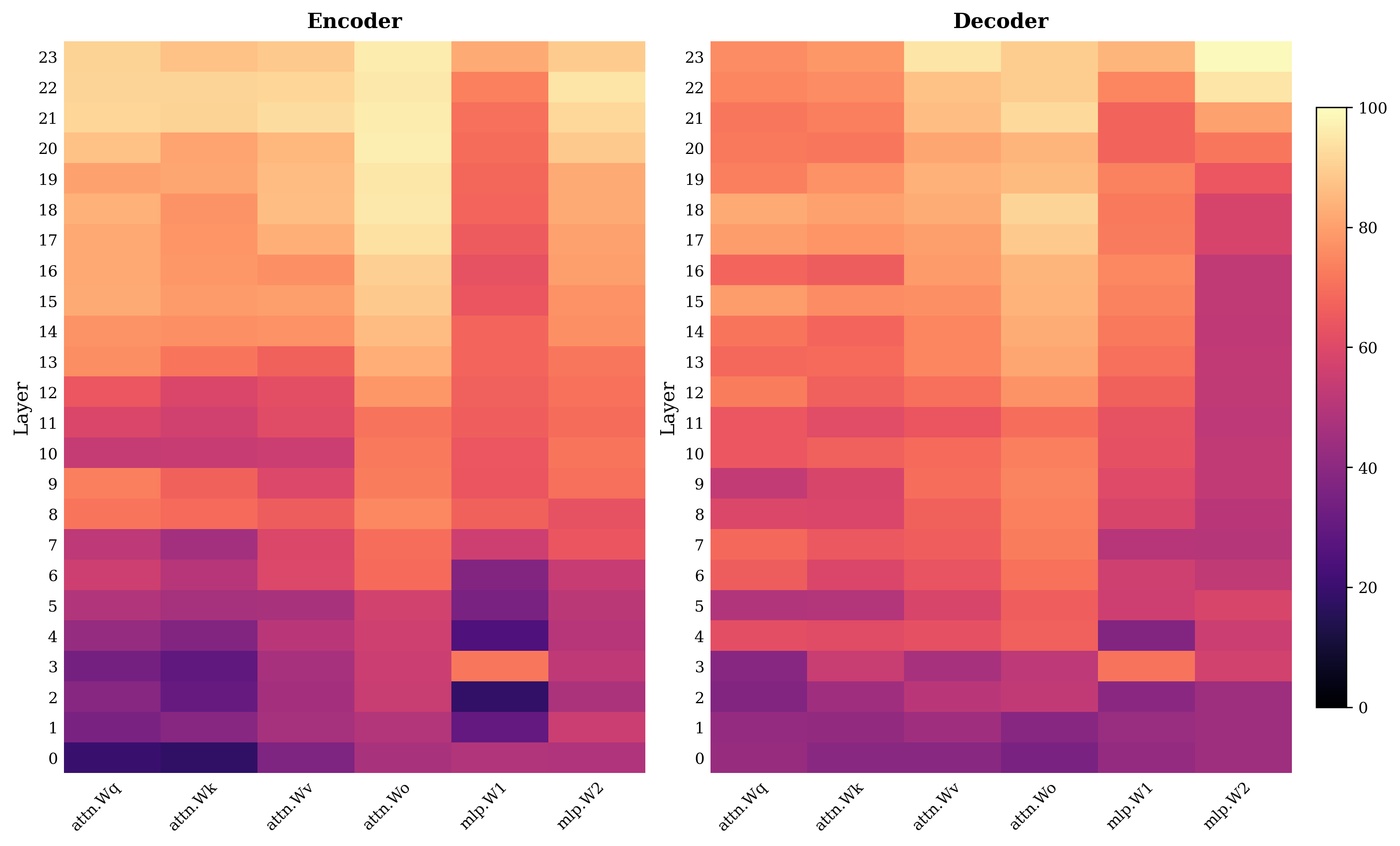}}
    \caption{
      Layer-wise row-level MAV threshold ratios between the pretrained \textsc{Whisper-medium} model and the \textsc{ja-en} code-switching fine-tuned model. 
    }
    \label{fig:mav_ja_en}
  \end{center}
\end{figure*}

\begin{figure*}[ht]
  \vskip 0.2in
  \begin{center}
    \centerline{\includegraphics[width=0.9\linewidth]{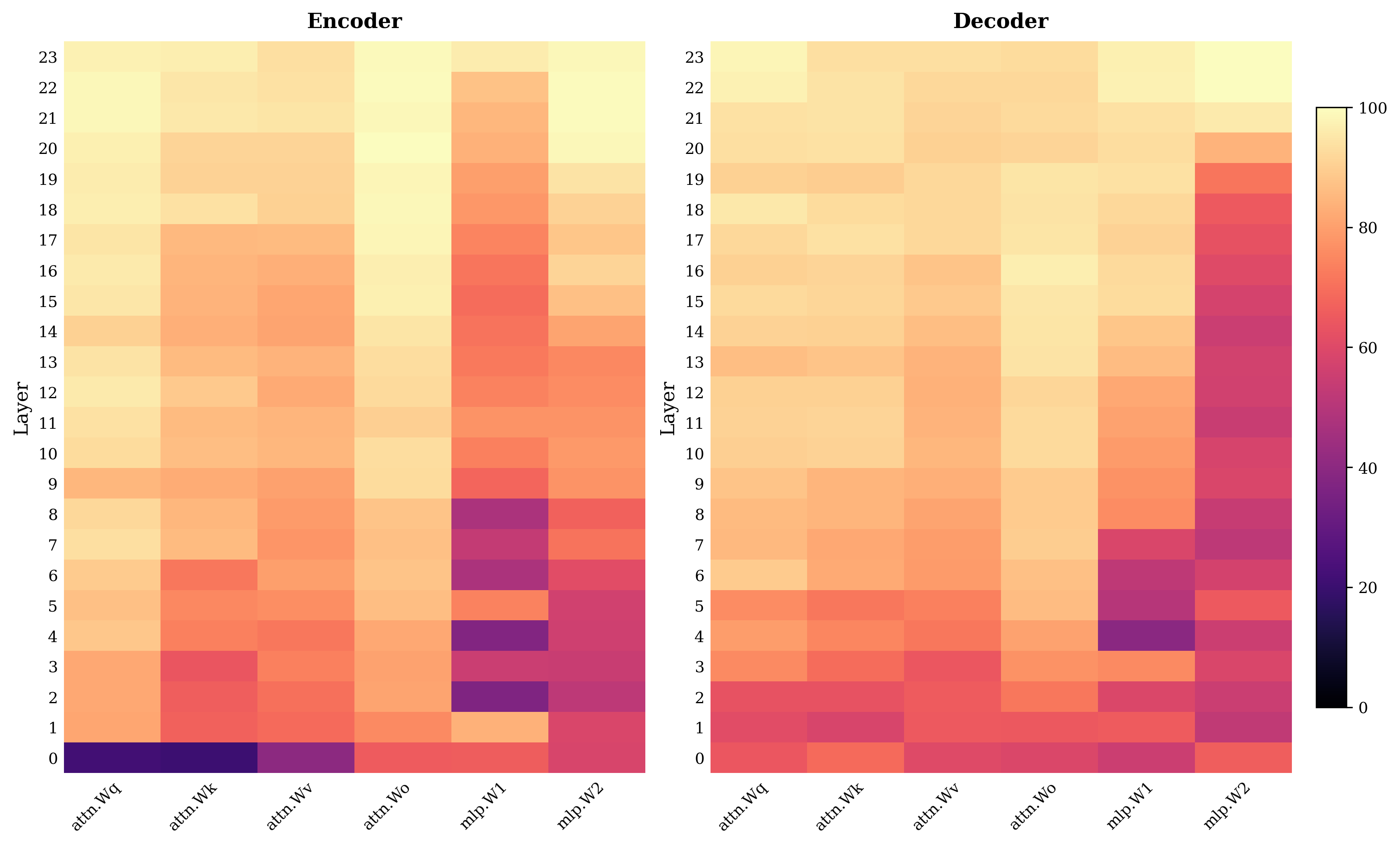}}
    \caption{
      Layer-wise row-level MAV threshold ratios between the pretrained \textsc{Whisper-medium} model and the \textsc{de-en} code-switching fine-tuned model. 
    }
    \label{fig:mav_de_en}
  \end{center}
\end{figure*}


\end{document}